\pdfoutput=1
\documentclass[11pt]{article}

\usepackage{acl}

\usepackage{times}
\usepackage{latexsym}
\usepackage{graphicx}

\usepackage[T1]{fontenc}
\usepackage[utf8]{inputenc}

\usepackage{microtype}

\title{Divide et Impera: \\Multi-Transformer Architectures for Complex NLP-Tasks}

  \author{
    Solveig Helland \\
    Department of Informatics\\ 
    University of Zurich \\
    Switzerland \\
    \texttt{solveig.helland@uzh.ch}
\And
    Elena Gavagnin \\
    Institute of Business Information Technology \\
    Centre for Artificial Intelligence \\
    Zurich University of Applied Sciences\\
    Switzerland \\
    \texttt{gava@zhaw.ch} \\
\AND    
    Alexandre de Spindler \\
    Institute of Business Information Technology \\
    Zurich University of Applied Sciences\\
     Switzerland \\
    \texttt{desa@zhaw.ch}   
}

\begin{document}
\maketitle
\begin{abstract}
The growing capabilities of transformer models pave the way for solving increasingly complex NLP tasks. A key to supporting application-specific requirements is the ability to fine-tune. However, compiling a fine-tuning dataset tailored to complex tasks is tedious and results in large datasets, limiting the ability to control transformer output.
We present an approach in which complex tasks are divided into simpler subtasks. Multiple transformer models are fine-tuned to one subtask each, and lined up to accomplish the complex task. This simplifies the compilation of fine-tuning datasets and increases overall controllability. Using the example of reducing gender bias as a complex task, we demonstrate our approach and show that it performs better than using a single model.
\end{abstract}

\section{Introduction}
Transformer models have received increased attention over the recent years. Much progress was achieved by improvements to model architectures, components, and algorithms such as from RNN to LSTM or GRU~\cite{ChungGCB14}, and from seq2seq~\cite{Sutskever2014} to attention~\cite{vaswaniAttentionAllYou2017,Bahdanau2015}, and GLM~2.0~\cite{Du2021} to name a few. Progress also resulted from vastly increasing parameters, such as GPT-2~\cite{Radford2019} with 1.5 billion, GPT-3~\cite{NEURIPS2020} with 175 billion, and Google Switch~\cite{Fedus2021} with 1.6 trillion parameters among others.

However, the training of a transformer model from scratch requires amounts of training data and computing power by far exceeding the scope of individual application development. Furthermore, while pre-trained models perform well when applying basic NLP tasks to common and broadly defined domains, they tend not to meet the requirements of more complex tasks applied to less common and more narrowly defined domains.

A key element supporting a wide variety of applications is the simplicity with which a pre-trained model may be turned into a special-purpose model by means of fine-tuning: the act of progressively adapting a subset of model weights based on a task- or domain-specific dataset.

For example, a domain-specific question answering (q\&a) model may be obtained using a fine-tuning dataset containing pairs of questions and answers related to that domain. However, while such a model may achieve an acceptable answering performance, it is unclear which discriminating features it is able to capture from questions, and what knowledge is applied when answers are generated.

This makes it difficult to create a q\&a model when particular requirements are imposed to answer content and wording. Furthermore, if question contexts such as topics of discourse, sentiments or user education should affect the answer content and wording, the fine-tuning dataset must include all combinations of context, content and wording. However, increasing the complexity of the dataset complicates the predictability of model responses.

In this paper we introduce a method for building transformer applications where model responses %must be 
are controllable while complex tasks are performed. We present the results of initial experiments conducted with GPT-3 to reduce gender bias in English texts. The results suggest further experiments with other complex tasks, which will be reported in a more comprehensive publication.

The rest of the paper is structured as follows. In Section~\ref{sec:background} we present some basic notions and challenges dealing with gender bias removal, which represents for us an exemplary task where to test our novel approach, introduced in Section~\ref{approach}. In Section~\ref{results} we compare our approach to two baseline architectures and we show the performance in terms of bias reduction. In Section~\ref{discussion} we discuss our preliminary results and envision further extension of the approach.

\section{Gender Bias Removal as NLP Task}\label{sec:background}
We introduce gender bias removal as an NLP task where multiple input features must be taken into account, while model responses must be controlled.

Previous work on gender bias focused on specific NLP aspects such as word embeddings~\cite{Bolukbasi2016, Zhao2018, Kaneko2019}, coreference resolutions~\cite{Zhao2018b}, and part-of-speech and dependency parsing~\cite{Garimella2019}. However, these approaches share several limitations regarding their effectiveness of removing bias in texts~\cite{dinanMultiDimensionalGenderBias2020, kanekoGenderpreservingDebiasingPretrained2019, ethayarajhUnderstandingUndesirableWord2019}.
First, they tend to conflate different conversational dimensions of gender bias and are thus unable to detect subtle pragmatic differences.
Second, they are often limited to explicitly binarly gendered words while many words are not explicitly gendered.
Third, focusing on the male-female gender direction they neglect the impact of words that have a gender orientation but are not necessarily unfairly biased.

A more general framework was proposed in~\cite{dinanMultiDimensionalGenderBias2020} where textual gender bias is decomposed along three different pragmatic and semantic dimensions, such as bias 1.~from the gender of the person being spoken about, 2.~from the gender of the person being spoken to, and 3.~from the gender of the speaker.
It was shown that the distinction of gender bias along multiple dimensions generates better and more fine-grained gender bias classifiers. Consequently, we define gender bias removal as a complex NLP, which can benefit from a multi-step approach, including bias detection, classification and reformulation.

In this paper we characterise the occurrence of gender bias in two aspects. One aspect is the identification of the bias type, such as the use of gender-exclusive keywords for a gender-neutral entity (explicit bias,  \cite{hittiProposedTaxonomyGender2019, leavyGenderBiasArtificial2018a}), or the reference to a gender-neutral term through a gendered pronoun (generalisation bias, \cite{hittiProposedTaxonomyGender2019}), or expressions of well meant attitudes towards one gender guided by stereotypes (benevolent sexism, \cite{glickAmbivalentSexismInventory1996}). The second aspect then captures the actual terms having such a type of bias. As a result of bias being characterised in two aspects, the treatment of text must be specific to their combination. For each pair of bias type and terms having this type of bias, an appropriate reformulation must be applied.

\section{Approach}\label{approach}

The key to our approach is to break down a complex task into simpler subtasks. For each subtask, a dataset is created, on which a task-specific transformer model is fine-tuned. Since a subtask is less complex, these task-specific datasets remain small, and the behaviour of each model is thus more controllable.  Once all models are fine-tuned, they can be lined up to complete the overall task.

\subsection{Debiasing}

We define the three subtasks \textit{bias classification}, \textit{bias extraction}, and \textit{text reformulation} for which dedicated transformer models are used. 

\begin{figure*}
\centering
\includegraphics[width=0.8\textwidth]{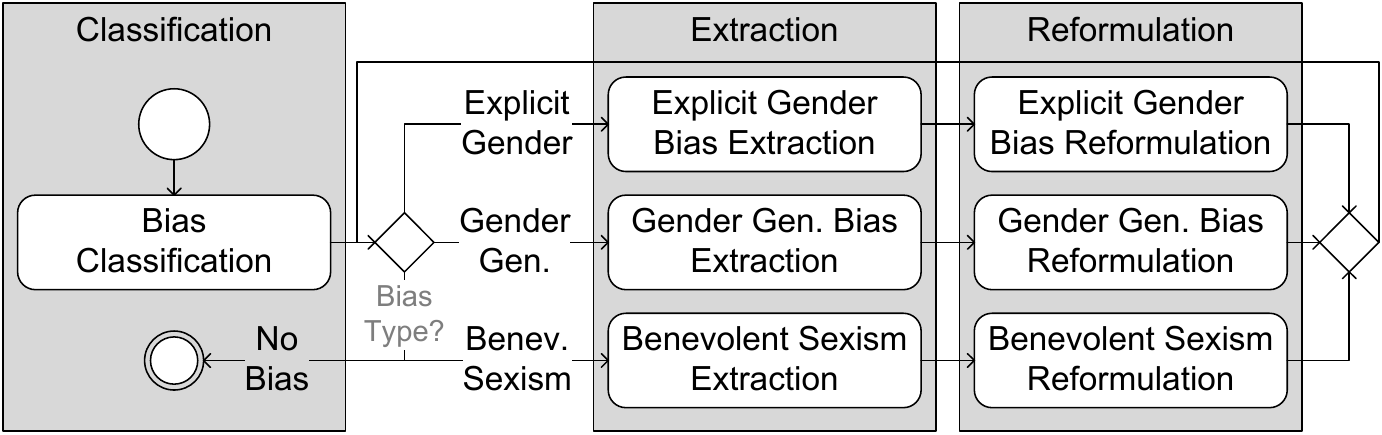}
\caption{Activity diagram of the debiasing approach with three subtasks (grey boxes) and seven transformer models (white boxes) in total.}
\label{fig:model3}
\end{figure*}

For the first subtask, a model is used to identify if and which type of bias a sentence has: gender generalization bias, explicit gender bias, benevolent sexism, or no bias. If no bias is detected, the debiasing process is halted.
The second subtask is to extract the terms that concern a bias. For each bias type there is a model able to perform a type-specific extraction. Thus, the bias type identified in the first subtask is used to select the appropriate model. 
As third subtask, the bias is removed using a model able to reformulate text. Similar to the second subtask, a model is available for each bias type. The type identified in the first subtask is thus used to select the appropriate model. The bias-carrying terms extracted in the second subtask are provided as input in addition to the text to be reformulated. 
Since a text may contain several biases of different types, an iterative approach is used, in which the text is repeatedly classified and treated until it results bias-free. Figure~\ref{fig:model3} shows the transformer architecture and debiasing process. 

All models are GPT-3 {\tt davinci} models, accessed through the OpenAI API. For details regarding GPT-3 architecture we refer to the original paper \citep{NEURIPS2020}. The temperature $T$ was set to $0.2$, $Top P$ to $1$, and $Best Of$ to $1$. The fine-tuning datasets for each subtask contained 10 input-output examples. The dataset for the classification model that performed the first subtask contained 10 examples of each bias type, yielding a total of 40 examples given in the prompt as fine-tuning dataset.

\subsection{Data}
To obtain the fine-tuning and test datasets, we first created a gender-bias dataset %bias-labeled dataset 
using Wikipedia's {\it neutral point of view} (NPOV) edits. An NPOV edit is a sentence that has been reported as biased and has therefore been changed by a Wikipedia contributor. Both versions of the sentence are stored in the revision history of the NPOV edits. %, and we have used 55'000 pairs. 
To focus on gender bias, sentences that did not contain gender pronouns were excluded, which determined a final size of 45'539 sentence-pairs (Wikipedia NPOV dataset). By comparing the biased and unbiased forms of the sentence, the bias-inducing terms and their unbiased substitutes were extracted.

Finally, a bias-labeled subset from the Wikipedia NPOV dataset was selected, in which all bias types were represented with equal frequency. In this subset, the biased sentences were manually labeled according to the three types of bias. Overall, the bias-labeled subset contained 40 examples to be used in the prompts for fine-tuning and 92 examples as test dataset.

The fine-tuning dataset for the first subtask (bias classification) was composed of biased sentences paired with the bias types. The three datasets for the second subtask (bias extraction) contained the biased sentences paired with the bias-inducing terms, grouped by the bias type. For the third subtask (text reformulation), three datasets were created where the biased sentences and the bias-inducing terms were paired with the unbiased sentences, grouped by bias type.

For the evaluation, 92 bias-labeled NPOV edits were used to create a test dataset containing the biased and unbiased sentences, the bias-inducing terms, and labeled with the bias types.

\subsection{Bias Measurement}\label{bias_measurement}
Two different approaches were developed to measure the bias before and after treatment.

The first approach validates each subtask using the F1 score. In the classification and extraction subtasks, this is done by comparing the model output to the expected output given in the test dataset. The reformulation subtask is validated by comparing reformulated sentences to the unbiased ones from the test dataset. Because of the diversity of how sentences can be rephrased, the bias classifier was again applied to the reformulated sentence. As a result, if a reformulated sentence did not match the unbiased sentence from the test dataset, we could still provide an indication of whether the bias had been removed.

The second approach uses GloVe word embeddings~\cite{Pennington2014} to quantify the result of debiasing. Following \cite{bolukbasiManComputerProgrammer2016}, we identified a gender direction (subspace) based on the gendered word pair {\it she-he}. This subspace is then used to evaluate the position of words that typically have a strong gender association in terms of gender direction.

For each word $w$, we computed the cosine similarity between its vector representation $\vec{w}$ and the vector representation of the gender pronouns $\vec{she}$ and $\vec{he}$. 
The degree of a word's gender-neutrality, called {\it word neutrality}, is defined as $\cos(\vec{w},\vec{she}) - \cos(\vec{w},\vec{he})$ where neutral words tend towards zero. Aggregating and normalizing over the whole vocabulary $W$  \[ \frac{1}{N_W} \sum_{w \in W}(\cos(\vec{w},\vec{she}) - \cos(\vec{w},\vec{he}))^2, \] with $N_W$ being the vocabulary size, we get an overall measure for gender bias in a text, namely the {\it mean squared word neutrality (MSWN)}.
% @Elena, you are the best! :-)

\section{Results and Evaluation}\label{results}

To evaluate our approach, we designed two additional transformer architectures as base lines. The first one (M-1) consists of a single transformer model fine-tuned to perform debiasing as one single task.
The second one (M-2) is a double-transformer system, composed by a first model identifying the bias type which selects the second model for type-specific reformulation.
In what follows, we refer to our approach as M-3. We debiased the 92 sentences from the test dataset using M-1, M-2, and M-3. We then used the bias measurements \textit{F1 Score} and \textit{Mean Squared Word Neutrality} introduced above to quantify each architecture's performance and the debiasing process.

\subsection{F1 Scores}

Debiasing is deemed successful when the bias-inducing terms are removed and replaced with unbiased alternatives, resulting in an unbiased sentence. The F1 scores are shown in Table~\ref{comparison_models}. In comparison, the benefits of each split in subtasks emerges clearly: 100\% improvement in micro averaged F1 score from M-1 to M-2, and an additional 50\% improvement from M-2 to M-3. 

\begin{table}
\centering
\begin{tabular}{lccc}
\hline \textbf{Gender Bias Type} & \textbf{M-1} & \textbf{M-2} & \textbf{M-3}\\ \hline
Benevolent sexism & 0.09& 0.57 & 0.87\\
Explicit gender bias & 0.18  & 0.27 & 0.95 \\
Gender generalization & 0.65 & 0.87 & 0.91 \\
\hline
Micro average & \textbf{0.31} & \textbf{0.57} & \textbf{0.91}\\
\hline
\end{tabular}
\caption{Comparison of debiasing F1 scores for the approch proposed in this paper (M-3) and the two baselines (M-1 and M-2).}
\label{comparison_models}
\end{table}

Interestingly, differences in F1-score occur also along different gender bias type. Apparently, some more subtle forms of gender bias require a more extensive treatment than others. This is the case for example of benevolent sexism or explicit gender bias, which in most of the cases fail to be corrected by using the base lines models. However, for the simpler case of gender generalization bias even a single transformer provide satisfactory results.

The introduction of the bias classification transformer in M-2 and M-3, improves the performance at reformulation stage, also due to the fact that treating each bias type separately, allows the prompt to contain more examples, providing therefore more fine-tuning training. 
Consequently, a multi-transformers approach presents the advantage to be more flexible and scalable across different complexity level within a task.

\subsection{Mean Squared Word Neutrality}

We computed the MSWN for the test dataset as well as for the complete Wikipedia NPOV dataset, see values in Table~\ref{debiasing}. We are thus able to compare the transformer-based approach to the work of the Wikipedia contributors. For comparability, we considered as vocabulary base the profession titles from~\cite{bolukbasiManComputerProgrammer2016}, and descriptive words  from~\cite{gaucherEvidenceThatGendered2011}. Of the original lists considered, 49 professions and 67 descriptions appear in the Wikipedia NPOV dataset, while 8 and 7 resp. are found in the test dataset.

An MSWN closer to zero means less gender-bias encountered. Independent of the set of words considered, the MSWNs get closer to zero after treatment. Moreover, this does not only apply to debiasing performed by humans (Wikipedia NPOV dataset) but also to the sentences from the test dataset which were debiased following our M-3 approach. 

\begin{table}
\centering
\begin{tabular}{lccc}
\hline \textbf{Dataset} & \textbf{Professions} & \textbf{Descriptions} \\ \hline
NPOV & 0.0090 & 0.0057 \\
NPOV Debiased & 0.0049 & 0.0040  \\
\hline
Test Set & 0.0207 & 0.0085  \\
Test Set Debiased & 0.0156 & 0.0065  \\
\hline
\end{tabular}
\caption{ MSWN in Wikipedia NPOV dataset (biased and debiased by humans), and in the test dataset (biased and debiased using M-3).}
\label{debiasing}
\end{table}

\section{Discussion and Conclusions}\label{discussion} 

In this paper we presented a new approach to address complex NLP tasks such as gender-bias removal. The multitude of aspects which characterize different type of gender bias proved to favour our iterative multi-step approach, where the ultimate task (bias removal) is split into simpler subtasks. Each subtask is performed by a dedicated, specifically fine-tuned transformer model. Our approach proved to be effective when the models were fine-tuned using a handful of sentences, in contrast to using a single model, which potentially need more fine-tuning data to provide comparable results.

The advantage presented from this task-splitting is not only to have each transformer do a simpler task, but also the possibility to generate more straightforward input-output data combinations for fine-tuning.  Moreover, our approach can be extended by including an arbitrary number of bias types and treatments.
% Alex: possibly dangerous statement? without making the process more complex and expensive

However, our approach currently treats multiple biases one by one, iteratively. This could be improved if the approach would be extended to treat multiple biases at once.

The results of applying our approach to bias removal in natural language texts indicated that the method proposed is effective. In order to explore its applicability in more general settings, we aim to apply it to other NLP tasks and use different transformer models.

\bibliographystyle{acl_natbib}
\bibliography{Solveig_final}

\end{document}